# Crowdsourcing Parallel Corpus for English – Oromo Neural Machine Translation using Community Engagement Platform

Sisay Chala[1], Bekele Debisa[2], Amante Diriba[3],
Silas Getachew[4], Chala Getu[5], Solomon Shiferaw[6]

[1]chala@fit.fraunhofer.de, [2]bdebisa@gmail.com, [3]bulidiriba@gmail.com,
[4]silogecho@gmail.com, [5]ntnlgetu@gmail.com, [6]s.seshiferaw@gmail.com

**Abstract**

Even though Afaan Oromo is the most widely spoken language in the Cushitic family by more than fifty million people in the Horn and East Africa, it is surprisingly resource-scarce from a technological point of view. The increasing amount of various useful documents written in English language brings to investigate the machine that can translate those documents and make it easily accessible for local language. The paper deals with implementing a translation of English to Afaan Oromo and vice versa using Neural Machine Translation. But the implementation is not very well explored due to the limited amount and diversity of the corpus. However, using a bilingual corpus of just over 40k sentence pairs we have collected, this study showed a promising result. About a quarter of this corpus is collected via Community Engagement Platform (CEP) that was implemented to enrich the parallel corpus through crowdsourcing translations.

## 1 Introduction

Machine translation (MT) is the process of translating text from one language (i.e., source language) to its corresponding text in another language (i.e., target language). Due to its practical demands, MT has been studied for quite many decades leading to various techniques with diverse and complex pros and cons, namely Rule-based Machine Translation (RBMT) and Corpus-based Machine Translation. Th former primarily utilizes rules developed by linguists whereas the latter is based on corpus of text out of which it generates patterns that help predict target text for the given source text.

Corpus-based MT is further divided into Statistical Machine Translation (SMT) and Neural Machine Translation (NMT). Associated with it are a number of challenging aspect aspects of MT with respect to its requirements and complex nature of human languages. The main challenge is the availability of comprehensive rules for RBMT and good quality corpus for SMT and NMT. There are also other challenges such as variety of languages, alphabets and grammatical structures; multiple possibilities of translation; and the natural evolution of languages.

In this study, we focus on developing English-Oromo MT using openNMT (Klein, Kim, Deng, Senellart, & Rush, 2017). We selected this approach because, unlike SMT which requires a pipeline of separate tasks, namely language modeling, translation modeling and decoding -- NMT is built with one network instead of a pipeline of separate tasks. Moreover, unlike RBMT, NMT does not require intensive linguistic preprocessing of the data.

Though NMT emerged in recent developments, its models outperformed the SMT (Bahdanau, Cho, & Bengio, 2014) leading many companies that offer MT services including Google (Wu, et al., 2016) and Microsoft to switch to NMT (Le & Schuster, 2016) (Microsoft Translator, 2016). Furthermore, as shown in **Figure 1**, the performance of neural MT consistently outperforms the other approaches (Le & Schuster, 2016) as tested for multiple language pairs.

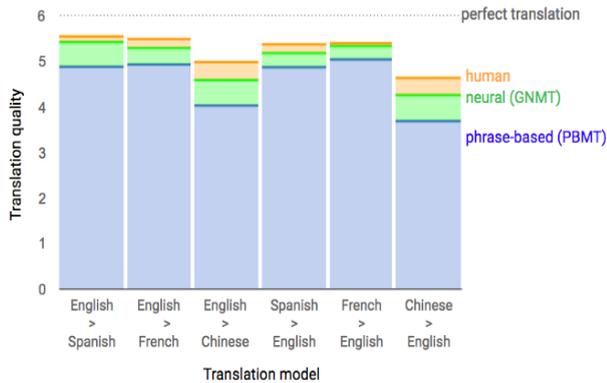

*Figure 1 Translation quality of statistical and neural MT models by Google (Le & Schuster, 2016)*

The paper is composed as follows: Section 2 discusses the analysis of related works providing a brief highlight of similar studies; Section 3 describes tools and methods utilized in the design and implementation of the system; Section 4 presents the prototype English-Oromo neural machine translation system; in Section 5 we discuss implemented measures to deal with the effect of limited linguistic resource; in Section 6 and 7 we discuss some results, and provide conclusions and possible future directions for research, respectively.

## 2   Analysis of Related Works

Google translate - the most popular translation platform supports over 100 languages at various levels and serves over 500 million people daily. In February 2016 Google announces adding 13 new languages including Ethiopia's Amharic which is the second most widely spoken Semitic language following Arabic, Xhosa of South Africa to its live translating tool, which is a breakthrough for our country's translation technology. Afaan Oromoo is not yet translated by Google.

All existing research on English to Afaan Oromoo translation uses either Statistical Machine Translation (SMT) or Rule Based Machine Translation (RBMT), both the above techniques have been replaced by different organization (including Google and Microsoft) by the new approach called neural machine translation. The methods have their own weakness in performing translation, NMT has advantage over both methods.

SMT, while translating, looks in millions of documents in order to find patterns to decide on the best translation. NMT translates "whole sentences at a time, rather than just piece by piece. It uses this broader context to help it figure out the most relevant translation, which it then rearranges and adjusts to be more like a human speaking with proper grammar" (Alimi & Amiri, 2018).

A review of articles to get an overview of translation score in relation to data size and quality for other languages is summarized as follows:

- English to Afaan Oromoo SMT (Adugna & Eisele, 2010) utilized bilingual corpora 20,000 sentences and monolingual corpora of 62,300 sentences, and achieved overall BLEU (Papineni, Roukos, Ward, & Zhu, 2002) score of 17.4%.

- Bidirectional English – Afaan Oromo Using Hybrid Approach by Jabesa Deba (Daba & Assabie, 2014) employed bilingual corpora of 3,000 sentences and achieved BLEU score of 37.41% and 52.02% for English to Afaan Oromo and for Afaan Oromoo to English, respectively.

- Bidirectional English-Amharic Machine Translation: An Experiment using Constrained Corpus by Elleni (Teshome, 2013) reported BLUE score 82.22% and 90.59%, for simple sentences English to Amharic and Amharic to English, respectively. Furthermore, for Complex sentence, the respective BLEU score for English to Amharic and Amharic to English was 73.38% and 84.12%.

- English-Afaan Oromo SMT by Yitayew Solomon and Million Meshesha (Meshesha & Solomon, 2018) utilized bilingual corpora of 6,400 sentences and monolingul corpora of 19,300 (for English) and 12,200 (for Afaan Oromoo), and reported BLEU score of 27% for phrase level with maximum phrase of length 16.

## 3  Tools and Methods

Even though there are number of research and studies about automatic translation of English to Afaan Oromoo, fully function platform where a user can easily translate text is still lacking. This study is set out to implement neural machine translation system in which we provide web-based translation.

While the main objective of this study is to demonstrate a prototype of English-Oromo neural machine translation system, the specific goals are threefold: the first is to develop automatic translation of text from English to Oromo using NMT, and the second is to compare the accuracy of NMT models with SMT models developed for English-Oromo language pairs (Adugna & Eisele, 2010). The third and final objective is to develop community engagement platform (CEP) in order to improve the accuracy of the baseline MT system by collecting more accurate parallel text through crowdsourcing (Nishimura, Sudoh, Neubig, & Nakamura, 2018).

## 4  Prototype English-Oromo NMT System

### 4.1  Data Collection and Preprocessing

To perform the experiments, the datasets or corpora were collected from documents from diverse sources such as Ethiopian criminal code, Ethiopian constitution, Oromia Regional State Duties and Responsibilities, Religious Book, books and article form Oromo medias, international conventions, translated document form Oromia Regional bureau, some translated books such as fictions, history, phycology. We also used a monolingual Afaan Oromo and English corpora collected from the web.

We performed data cleaning and preprocessing to make the dataset ready for alignment and experimentation. In other words, whenever the data is gathered from different sources it is collected in raw format which is not feasible for the analysis. So, we need data preprocessing to achieve better result from applied model in NMT. Data preprocessing includes data cleaning (removing unnecessary characters that affect translation quality), data integration (aggregating data obtained from different sources), data reduction (e.g., eliminating part of the data that does not have equivalent translation) and data transformation. For this purpose, we used python programming language.

### 4.2  Description of dataset

In order to train the NMT model we just need two files: source language file and target language file. Each with one sentence per line with words space separated. The standard format used in both statistical and neural translation is parallel text format, which consists of sentence pairs in plain text files corresponding to source sentences and target translations, aligned line-by-line and separated by space, following Chansung (2018).

### 4.3  Implementation of NMT System

Neural Machine Translation (NMT) is an end-to-end learning approach for automated translation, with the potential to overcome many of the weaknesses of conventional phrase-based translation systems. NMT is based on the model of neural networks in the human brain, with information being sent to different 'layers' to be processed before output. It makes faster translation than statistical method and has the ability to create higher quality output. NMT is also able to use algorithms to learn linguistic rules on its own from statistical models. The main strength of NMT is that it is able to learn directly, the mapping from input text to the corresponding output text in an end-to-end fashion.

Another major success of NMT "as Google has seen in its use of NMT, the technology has several advantages, including its application of a singular system that can be trained directly on both source and target text" (Ulatus, 2018). Another significant element of NMT is its capability to automatically fix its parameters throughout its training period. Other benefits include that NMT:

- Efficient translation of grammatically complex languages, such as Korean, Japanese, and Arabic (Ulatus, 2018).

- Uses algorithms to detect and learn language conventions that come from statistical models, resulting in quicker and better translations (Ulatus, 2018).
- Considers the complete sentence, not just a string of words.
- Learns nuances of languages, such as genders, inflection, and formality.
- Assists in applications, such as multilingual authoring, translation checking, and multilingual video conferencing (Ulatus, 2018).

We use OpenNMT which is open source initiative for neural machine translation and neural sequence. OpenNMT has the Following features such as simple general-purpose interface, requiring only source/target files, highly configurable models and training procedures. Recent research features to improve system performance and engages community from both academic and industrial contributions and requests.

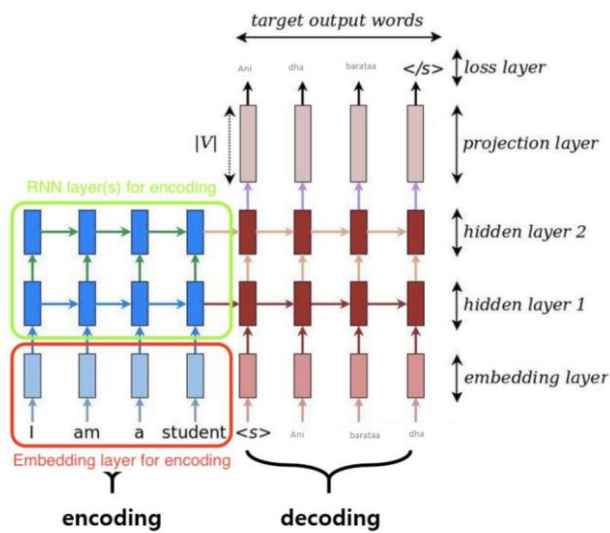

*Figure 2 Neural Machine Translation [Adapted from Park Chansung (2018) (Chansung, 2018)*

### 4.4 Evaluation

Machine translation systems can be evaluated by human or automatic evaluation method. Even if human evaluation is accurate it is costly and suffers from limited efficiency as compared to automatic evaluation. Therefore, we used BLEU score to evaluate the performance of the system, which is an automatic evaluation technique.

## 5 Handling Limited Language Resources

We implemented various techniques to deal with the effect of limited linguistic resources for English-Oromo language pairs on the effectiveness of the prototype machine translation. First, we tried to improve the translation quality by improving alignment of the sentences than the documents as in the baseline. Second, we implemented a Community Engagement Platform (CEP) in order to get more data of better quality from the community who provided us with data on voluntary basis.

### 5.1 Community engagement platform (CEP)

Community engagement platform is composed of two actions the community will take to contribute to the system, i.e., translation and verification as well as mechanisms for rewarding contributors in the form of gamification as described in the following subsections:

#### 5.1.1 Translation

Volunteers have contributed to the enrichment of our system by providing translations of the text on our platform (Lingogrid.com, 2020) sentence by sentence. One can provide a single translation or multiple translations as shown in Figure 3 and Figure 4, respectively and submit the translation sentence(s). One can also skip translating the sentence and proceed to the next sentence.

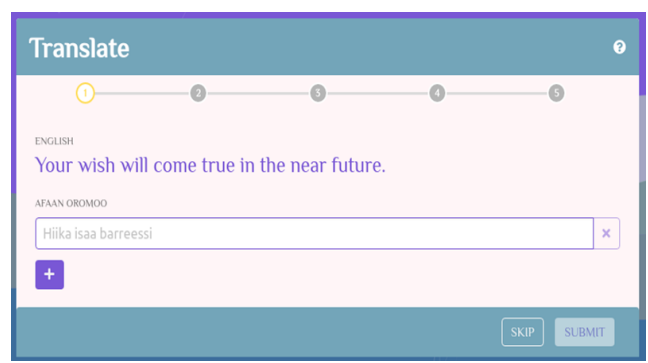

*Figure 3 Single translation of a given sentence*

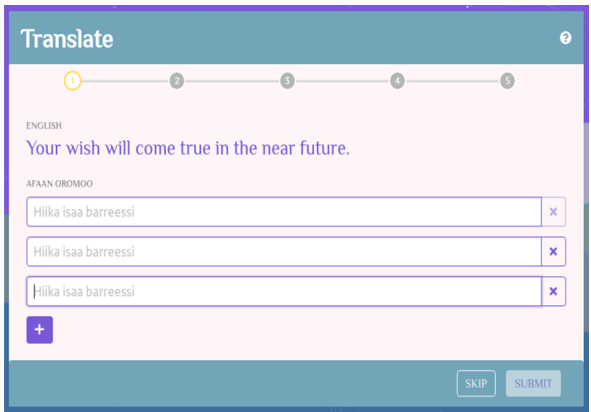

*Figure 4 Multiple translation of a given sentence*

### 5.1.2 Verification

In order to improve the quality of translations, users have also provided their contributions through verification (Lingogrid.com, 2020) of existing translations through an interface developed for this purpose. This enables us to validate translated texts generated automatically from parallel documents as well as those translations provided by volunteers.

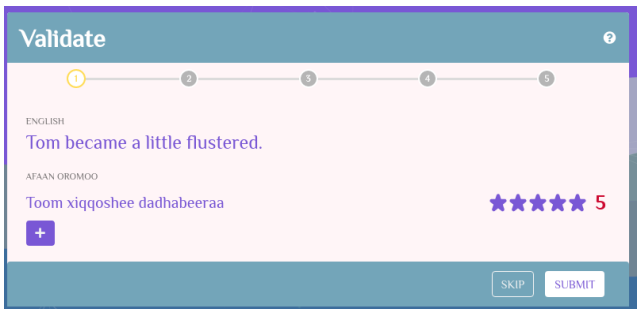

*Figure 5 Validating a translation with rating*

Users contribution on verification can be in two forms: i) rating the given translation as shown in Figure 5 and ii) rating and also providing alternative translation as shown in Figure 6.

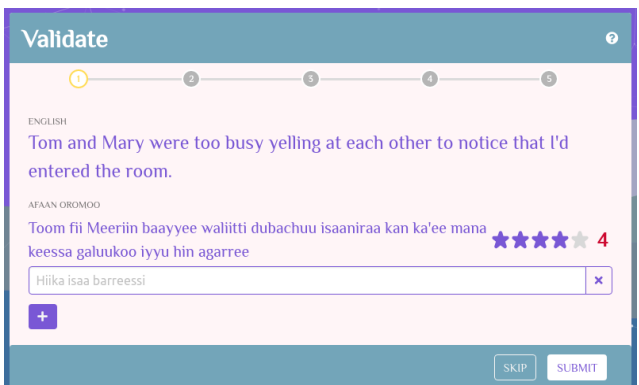

*Figure 6 Validating a translation with rating and alternative translation*

Both translation and verification are done in batches in order to not overwhelm the contributor and divide the contribution in manageable episodes of 5 contributions at a time as shown in Figure 7. The choice of 5 is arbitrary with the objective of gathering more inputs without imposing stress on users.

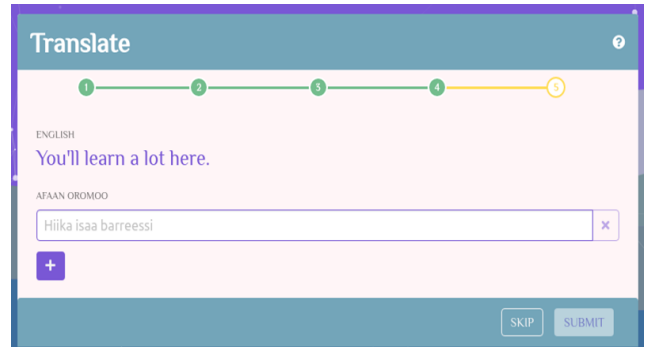

*Figure 7 Contributions in a batch of 5*

### 5.1.3 Gamification

Engaging users through gamification (Lingogrid.com, 2020), we have developed a rewarding system that can motivate individual contributors through badges and leaderboard as shown in Figure 8 and Figure 9 , respectively.

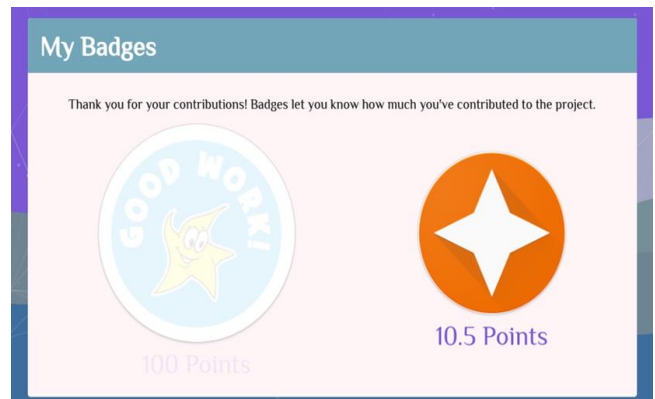

*Figure 8 Gamification with Badges*

While badges motivated individual contributors through sense of achievement, leaderboard motivated individual contributors through awareness of competitiveness.

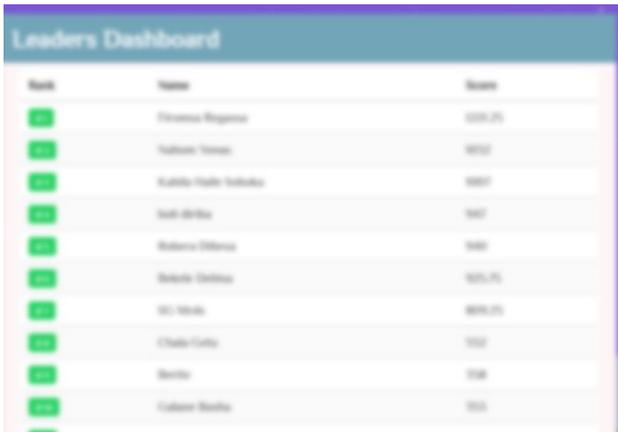

*Figure 9 Gamification with leaderboard (blurred to ensure privacy of the data providers)*

## 6 Results and Discussion

The main result of this study is to demonstrate the prototype system (Lingogrid.com, 2020) having interface shown in Figure 10.

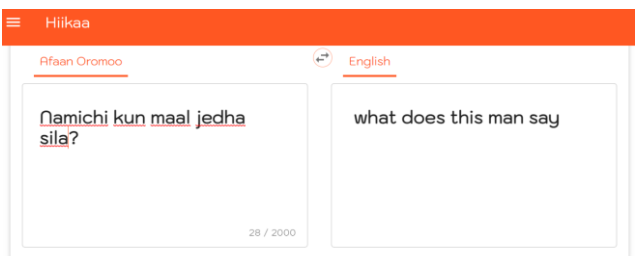

*Figure 10 Prototype English-Oromo Translation System (screenshot from (Lingogrid.com, 2020))*

Using the CEP, we were able to collect over 10k sentence pairs, the quality of which we improved through the crowdsourced verification.

Measuring quality of translation was done using BLEU score to try and guess automatically to simulate evaluation of human translator about the quality of a given machine translation. In this experiment, with 42k sentence pairs of bilingual corpus, the achieved BLEU scores of translating from English-to-Oromo and Oromo-to-English are 26% and 22%, respectively.

## 7 Conclusions and Future Works

We have presented the implementation of a prototype English-Oromo neural machine translation. In addition, aimed at increasing parallel corpus for under-resourced languages, we have presented a community engagement platform that was augmented with gamification and that enables collection of translations as well as validation of translated texts.

The future works on this study remain to find out the limitations of translation quality through more and more data of parallel text with high quality. One can also analyze cross-domain translation performance.

In addition, the usability of the CEP system from the users' point of view needs to be studied and quantified in order to identify points of improvement in look-and-feel as well as performance (i.e., response time, transaction rates and throughput). Expanding the MT system as well as the CEP for more language pairs is also another major issue for future work.

Another aspect of the future work will be to capitalize on machine translation outputs in supporting public administrations in order to provide inter-translation of content from one language to another. This will be of paramount importance in a multilinguistic communication setting.